\documentclass[11pt]{article}

\usepackage[margin=1in]{geometry}
\usepackage{fancyhdr}
\usepackage{titlesec}
\usepackage{tabularx}  % to auto-wrap long email
\usepackage{titling}
\usepackage{lmodern}
\usepackage{parskip}
\usepackage[colorlinks=false,linkcolor=blue,citecolor=blu]{hyperref}
\usepackage{url}
\usepackage{enumitem} % Required for customizing lists
\usepackage{graphicx}
\usepackage{soul}
\usepackage{cleveref}
\usepackage{float}
\usepackage{booktabs}
\usepackage{color}
\usepackage[square,numbers]{natbib}
\usepackage{changepage}  % for adjustwidth
\usepackage{setspace}
\usepackage{times}         % Times New Roman
\usepackage{changepage}    % For indented abstract
\usepackage{lmodern}       % Optional: for better spacing
\usepackage{hyperref}
\usepackage{etoolbox}  % For patching
\fancypagestyle{firstpage}{
  \fancyhf{}
  \fancyhead[L]{Accepted at the AMIA Annual Symposium 2025. The final version will appear in the official proceedings.}
  \fancyfoot[C]{\thepage}
  
}
\begin{document}
\thispagestyle{firstpage}
% \vspace{1em}
\noindent\rule{\textwidth}{4pt}
\vspace{1.5ex}

\begin{center}
  {\LARGE\bfseries Failure Modes of Time Series Interpretability Algorithms for Critical Care Applications and Potential Solutions}
\end{center}

% Bottom thin rule
\vspace{1.5ex}
\noindent\rule{\textwidth}{1pt}
\vspace{2em}
% \maketitle

% Author block
\begin{center}
  \begin{tabular}{c@{\hspace{2cm}}c}
    \textbf{Shashank Yadav} & \textbf{Vignesh Subbian} \\
    Department of Biomedical Engineering & Department of Biomedical Engineering \\
    University of Arizona & University of Arizona \\
    \texttt{shashank@arizona.edu} & \texttt{vsubbian@arizona.edu}
  \end{tabular}
\end{center}

% \vspace{2em}

% \begin{abstract}

\vspace*{22pt}  % Two lines of spacing before abstract (2 × 11pt)

\begin{center}
  \textbf{\fontsize{12pt}{12pt}\selectfont Abstract}
\end{center}

\begingroup
\fontsize{10pt}{11pt}\selectfont
\begin{adjustwidth}{0.5in}{0.5in}  % Indent both sides by 0.5 inch
Interpretability plays a vital role in aligning and deploying deep learning models in critical care, especially in constantly evolving conditions that influence patient survival. However, common interpretability algorithms face unique challenges when applied to dynamic prediction tasks, where patient trajectories evolve over time. Gradient, Occlusion, and Permutation-based methods often struggle with time-varying target dependency and temporal smoothness. This work systematically analyzes these failure modes and supports learnable mask-based interpretability frameworks as alternatives, which can incorporate temporal continuity and label consistency constraints to learn feature importance over time. Here, we propose that learnable mask-based approaches for dynamic time-series prediction problems provide more reliable and consistent interpretations for applications in critical care and similar domains.
\end{adjustwidth}
\endgroup
% \noindent\textbf{Abstract.} 

% \end{abstract}

% \section*{Abstract}
% \vspace*{-\baselineskip}
% \textit{Interpretability plays a vital role in aligning and deploying deep learning models in critical care, especially in constantly evolving conditions that influence patient survival. However, common interpretability algorithms face unique challenges when applied to dynamic prediction tasks, where patient trajectories evolve over time. Gradient, Occlusion, and Permutation-based methods often struggle with time-varying target dependency and temporal smoothness. This work systematically analyzes these failure modes and supports learnable mask-based interpretability frameworks as alternatives, which can incorporate temporal continuity and label consistency constraints to learn feature importance over time. Here, we propose that learnable mask-based approaches for dynamic time-series prediction problems provide more reliable and consistent interpretations for applications in critical care and similar domains.}

\section{Introduction}
% \vspace*{-\baselineskip}
% \st{Deep learning models are integral to healthcare analytics to predict patient outcomes, detect abnormalities, and support clinical decision-making 
% % \citep{begoli2019need}. 
% % However, due to high-stakes decision-making, simply providing accurate predictions is insufficient. Clinicians and other healthcare providers need to know that the model’s recommendations are based on reliable clinical reasoning \cite{dey2022human,stiglic2020interpretability}. 
% Models must offer clear and understandable explanations that experts can evaluate. As a result, the demand for interpretability in clinical applications is increasing, directly influencing user trust, patient outcomes, and patient safety 
% % \cite{rudin2019stop}. 
% }

% \textcolor{red}{
Interpretability techniques are essential in high-stakes, resource-constrained environments such as critical care medicine to ensure model-derived interpretations align with temporal dynamics of clinical trajectories. While deep learning models detect subtle temporal patterns, their clinical utility depends on interpretations that map predictions to pathophysiology\cite{dey2022human}. Time-series interpretability techniques must not only attribute feature importance but also contextualize when and why specific features matter, such as a blood pressure drop preceding cardiac arrest.  Conventional interpretability approaches may oversimplify the temporal granularity required in critical care, attributing importance to isolated time points rather than evolving physiological contexts. This mismatch risks confusing treatment artifacts with genuine deterioration in patient health\cite{stiglic2020interpretability}, highlighting the need for temporally coherent interpretability methods. Among conventional methods, Integrated Gradients\cite{sundararajan2017axiomatic}, DeepLIFT\cite{shrikumar2017learning} and GradientSHAP\cite{lundberg2018explainable} are foundational approaches that leverage gradients to attribute importance to features. However, gradient-based methods often struggle to capture temporal dependencies inherent in sequential data\cite{srinivas2020rethinking}, leading to adaptations like Temporal Integrated Gradients (TIG)\cite{enguehard2023time} and Sequential Integrated Gradients\cite{enguehard2023sequential}. 

Complementing gradient-based methods are perturbation-based approaches, which assess feature importance by altering parts of the input and observing the corresponding impact on predictions. Feature Occlusion (FO)\cite{suresh2017clinical} and Augmented Temporal FO\cite{tonekaboni2020went} are perturbation-based methods that rely on non-learnable input masking strategies or feature removal without dynamically adjusting to the data. In contrast, learnable mask-based methods such as DynaMask\cite{crabbe2021explaining} and ExtremalMask\cite{enguehard2023learning} optimize the masking process to either preserve or perturb key features while optimizing temporal continuity and label consistency constraints. Feature Importance in Time (FIT)\cite{tonekaboni2020went}, a method specifically developed for time-series interpretability offers a robust framework for quantifying shifts in predictive distributions and understanding temporal feature importance. Furthermore, certain model architectures enable inherent interpretability by identifying feature saliency, such as RETAIN\cite{choi2016retain}, by directly learning attention weights to provide feature attributions. However, reliance on attention mechanisms can lead to inconsistencies in interpretations as supported by natural language processing literature\cite{jain2019attention}. Existing methods for evaluating time-series interpretability have focused on static classification tasks, where a prediction is made only after observing the entire sequence, typically resulting in binary or multi-class output (e.g., EEG Classification, Patient Mortality Prediction). In contrast, this work extends time-series interpretability methods to dynamic prediction tasks, such as predicting acute organ failure, where predictions are generated at each time step, enabling insights into varying patient states during their stay in Intensive Care Units (ICU). In this paper, we contribute:

\begin{enumerate}[label=\arabic*., topsep=0pt, leftmargin=1.5em, labelwidth=1.2em, align=left]
    \item A systematic analysis of failure modes in time-series interpretability algorithms for dynamic organ failure prediction tasks, focusing on target selection, attribution aggregation, and temporal smoothness.
    
    \item Empirical evidence supporting learnable mask-based frameworks to address the failure modes through optimization with time-series specific constraints for dynamic organ failure prediction tasks.
\end{enumerate}

% \textbf{Note to VS: here we have to justify: This workshop invites submissions on negative results, failed experiments, and challenges in applying deep learning to real-world problems across various domains. It aims to foster open discussion, share insights to prevent repeated mistakes and promote transparency in the field. Papers should include a use case, proposed solution, negative outcomes, and an investigation of the reasons for failure, such as data issues, model limitations, or deployment challenges. Submissions will be evaluated based on rigor, novelty, insights, reproducibility, and clarity.}

\section{Methods}

\subsection{Dataset Description and Prediction Task}
% \vspace*{-\baselineskip}
We focused on Circulatory Failure, a leading cause of morbidity and mortality in critical care settings\cite{members2023heart}, as the use case for this work. We utilized Dynamic Circulatory Failure (refer to footnote\footnote{Circulatory failure is defined as lactate levels exceeding 2 mmol / L combined with mean arterial blood pressure below 65 mmHg or administration of any vasoactive drug.})  Prediction task from the HiRID-ICU benchmark study\cite{yeche2021hirid}, which involves dynamic binary prediction throughout the patient's ICU stay. Specifically, it continuously predicts the onset of circulatory failure within the next 12 hours, provided the patient is not already in organ failure.  Table \ref{table1} summarizes the distribution of the dataset for the dynamic circulatory failure prediction task, including the number of ICU stay records in the training, validation, and test sets, and the number of prediction samples. This multivariate time-series dataset spans 2016 time steps at 5-minute intervals, totaling 7 days, and includes 231 clinical features such as vital signs, hemodynamic data, treatments, pathological laboratory values, and ventilation parameters for critical care management. \\

% \begin{table}[H]%[t]
% \caption{Dataset Description for the Dynamic Circulatory Failure Prediction Task. M: Million}
% \label{tablea1}
% \begin{center}
% \begin{tabular}{ccc}
% \multicolumn{1}{c}{\bf Set} & \multicolumn{1}{c}{\bf ICU Stays} & \multicolumn{1}{c}{\bf Predictions(\% positive)} \\ \hline \\
% Train       &   23643        &   11.56M (4.51\%)          \\
% Validation  &   5072         &   2.42M  (4.22\%)          \\
% Test        &   5069         &   2.44M  (4.67\%)          \\
% \bottomrule
% \end{tabular}
% \end{center}
% \end{table}

\begin{table}[H]
\caption{Dataset Description for the Dynamic Circulatory Failure Prediction Task. M: Million}
\label{table1}
\centering
\begin{tabular}{ccc}
\toprule
\textbf{Set} & \textbf{ICU Stays} & \textbf{Predictions (\% positive)} \\ 
\midrule
Train       & 23643        & 11.56M (4.51\%) \\
Validation  & 5072         & 2.42M (4.22\%) \\
Test        & 5069         & 2.44M (4.67\%) \\
\bottomrule
\end{tabular}
\end{table}
% \vspace{-10pt} % Adjust the value as needed
\subsection{Model Architectures and Training}
% \vspace*{-\baselineskip}
We replicated the results of the HiRID-ICU Benchmark study\cite{yeche2021hirid}, which used the standard transformer encoder architecture (1.64 Million parameters)\cite{vaswani2017attention} with causal attention as the best performing model for dynamic circulatory failure prediction. We significantly improve their approach by utilizing a causal encoder-only variant of the CrossFormer model (28.6 Thousand parameters)\cite{zhang2023crossformer}. This choice was motivated by CrossFormer's state-of-the-art performance on time-series forecasting tasks and parameter efficiency. The reduced parameter count not only aligns with the need for efficient gradient calculations but also enables faster computation in the case of the perturbation-based method where multiple forward passes of the model are required. For a causal variant of the crossformer model, we adopted a causal attention-based modeling approach because past timesteps must not be influenced by future timesteps in the clinical time series analysis. We modified the CrossFormer-Encoder model by introducing a causal attention mask, which ensured that predictions at any given timestep are influenced only by current and previous timesteps. We implemented the causal attention mechanism by applying a triangular mask that sets the attention scores for future timesteps to negative infinity before applying the softmax operation. As a result, predictions at each timestep strictly adheres to temporal causality, ensuring that predictions at each timestep depend only on information available at or before that specific timestep, thus strictly preventing leakage of future data into the past.

\subsection{Model Interpretability Methods}
% \vspace*{-\baselineskip}
We applied 14 time-series interpretability methods to the dynamic circulatory failure prediction task and extracted corresponding attribution maps using \texttt{captum}\cite{kokhlikyan2020captum} and \texttt{time-interpret}\cite{enguehard2023time} Python libraries. A complete list of methods is provided in Table \ref{table2}. We selected two patient records, denoted as Patient A and Patient B hereafter, whose complete 7-day data was available and contained more than one period of circulatory failure.  This selection aimed to assess whether the observed failure modes of interpretability methods were consistent across different cases rather than being isolated occurrences. We demonstrate that the limitations of interpretability methods persist across different examples and highlight systemic failure modes in their application to dynamic prediction tasks.

For models producing output in the format \( T \times T \times F \times C \), we chose two consecutive time steps during which patients had circulatory failure ($C=1$) and extracted the corresponding attribution maps for each $T$. For the Temporal Integrated Gradients method, the failure class was set as $C=1$, as it represents circulatory failure. As a result, the attribution set for this method was reduced to \( T \times F \times 1\), ensuring that the analysis focused specifically on the contributions leading to failure. For models that generated attribution maps as \( T \times F \), we calculated the attribution maps across the entire sequence since the causal prediction model ensured that temporal causality was maintained. \\

\begin{table}[H]
\centering
\caption{Comparison of Model Interpretability Methods. \(T\) denotes the number of time steps, \(F\) denotes the number of features, \(C\) denotes the number of classes or outputs. Output attributions are extracted for a single ICU stay.}
\label{table2}
\begin{tabular}{clcc}
\toprule
\bf Index & \bf Method & \bf Library & \bf Output Attribution Shape \\
\midrule
1  & GradientSHAP                    & \texttt{captum}         & \(T \times T \times F \times C\) \\
2  & DeepLift                        & \texttt{captum}         & \(T \times T \times F \times C\) \\
3  & DeepLiftSHAP                    & \texttt{captum}         & \(T \times T \times F \times C\) \\
4  & Integrated Gradients            & \texttt{captum}         & \(T \times T \times F \times C\) \\
5  & Temporal Integrated Gradients   & \texttt{time-interpret} & \(1 \times T \times F \times C\) \\
6  & Sequential Integrated Gradients & \texttt{time-interpret} & \(T \times T \times F \times C\) \\
7  & Occlusion                       & \texttt{time-interpret} & \(T \times T \times F \times C\) \\
8  & Augmented Occlusion             & \texttt{time-interpret} & \(T \times T \times F \times C\) \\
9  & Feature Ablation                & \texttt{captum}         & \(T \times T \times F \times C\) \\
10 & Feature Permutation             & \texttt{captum}         & \(T \times T \times F \times C\) \\
11 & RETAIN                          & \texttt{time-interpret} & \(T \times F\) \\
12 & FIT                             & \texttt{time-interpret} & \(T \times F\) \\
13 & DynaMask                        & \texttt{time-interpret} & \(T \times F\) \\
14 & Extremal Mask                   & \texttt{time-interpret} & \(T \times F\) \\
\bottomrule
\end{tabular}
\end{table}

% \vspace{-10pt} % Adjust the value as needed
\section{Results}

\subsection{Model Performance Comparison}
% \vspace*{-\baselineskip}
The causal CrossFormer-Encoder model significantly outperformed the standard causal transformer model on the dynamic circulatory failure task. Specifically, our causal CrossFormer achieved higher performance across AUCROC, AUCPR, F1 and MCC score (Table \ref{table3}).  This improvement highlights the effectiveness of CrossFormer’s segment-based embedding and causal cross-dimensional attention in capturing temporal dependencies. Moreover, recent research has indicated that models that achieve higher performance provide better interpretations—providing clearer insights into model behavior \citep{liu2020impact, li2022interpretable}. Hence, the causal CrossFormer not only minimizes generalization error but also improves the quality of downstream interpretations.\\

% \begin{table}[H]%[t]
% \caption{Model evaluation metrics on the test set. Mean and standard deviation are calculated over ten runs. M: Million, K: Thousand}
% \label{table3}
% \begin{center}
% \begin{tabular}{ccccc}
% \multicolumn{1}{c}{\bf Model (\# Parameters)} & \multicolumn{1}{c}{\bf AUC-ROC} & \multicolumn{1}{c}{\bf AUC-PR} & \multicolumn{1}{c}{\bf F1} & \multicolumn{1}{c}{\bf MCC} 
% \\ \hline \\
% Causal Transformer-Encoder (1.64M)  &   90.26$\pm$0.42 &  34.84$\pm$0.69 &  26.32$\pm$2.56 &  28.72$\pm1.62$ \\
% Causal Crossformer-Encoder (28.6K)  &   97.19$\pm$0.20 &  68.05$\pm$0.52 &  59.87$\pm$0.71 &  58.75$\pm0.65$ \\
% \end{tabular}
% \end{center}
% \end{table}

\begin{table}[H]
\caption{Model evaluation metrics on the test set. Mean and standard deviation are averaged over ten runs. M: Million, K: Thousand}
\label{table3}
\centering
\begin{tabular}{lcccc}
\toprule
\textbf{Model (\# Parameters)} & \textbf{AUC-ROC} & \textbf{AUC-PR} & \textbf{F1} & \textbf{MCC} \\
\midrule
Causal Transformer-Encoder (1.64M)  & 90.26$\pm$0.42 & 34.84$\pm$0.69 & 26.32$\pm$2.56 & 28.72$\pm$1.62 \\
Causal Crossformer-Encoder (28.6K)   & 97.19$\pm$0.20 & 68.05$\pm$0.52 & 59.87$\pm$0.71 & 58.75$\pm$0.65 \\
\bottomrule
\end{tabular}
\end{table}

% \vspace{-10pt} % Adjust the value as needed
\subsection{Failure Mode 1: Time-Varying Multi-Output Models}
\begin{figure}[H]
\begin{center}
\includegraphics[width=\linewidth]{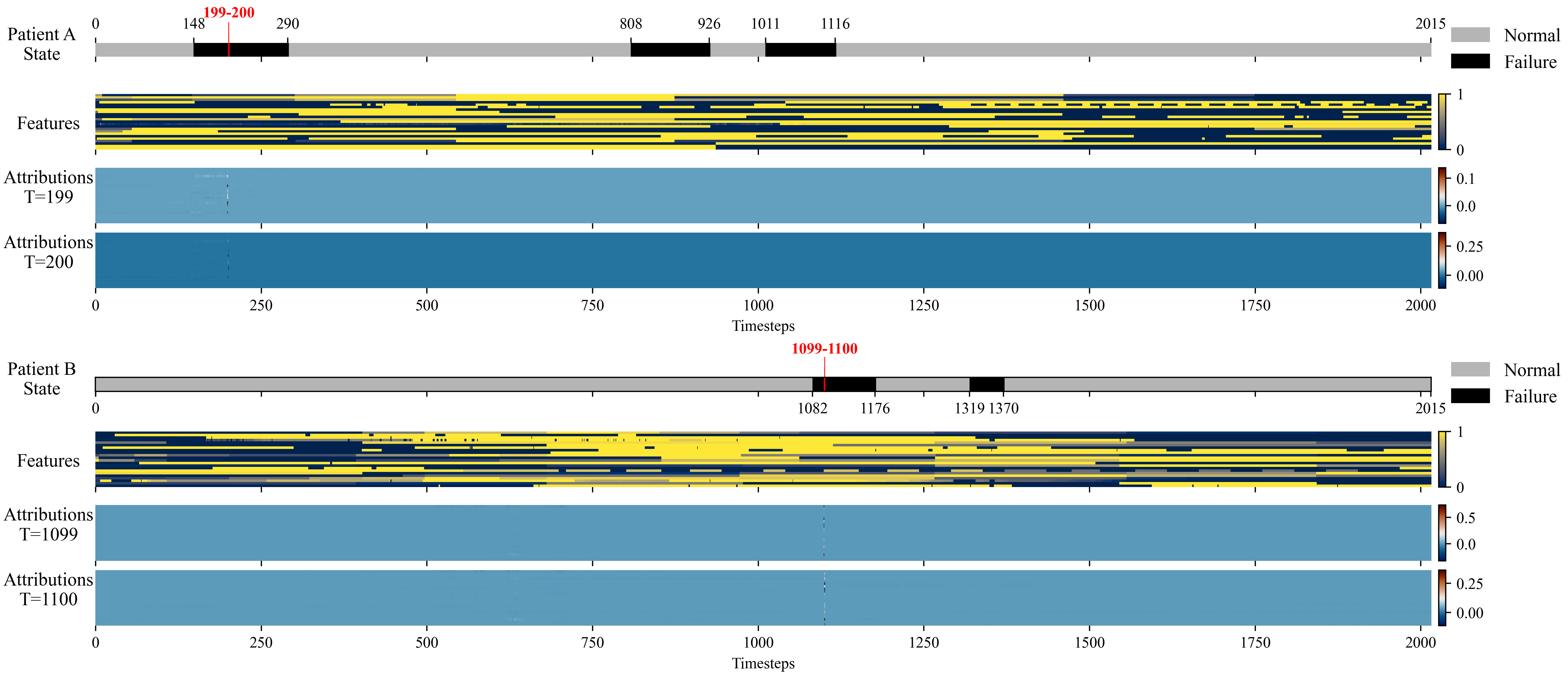}
\end{center}
\caption{Illustration of the timeline for Patient A and Patient B, from T=0 to T=2015, for the dynamic prediction task of circulatory failure. For each patient, the top bar indicates the patient's state over time, while the first heatmap displays normalized feature values across the timeline. Additionally, feature-time attribution heatmaps generated using the \textbf{GradientSHAP} method are presented for specific time steps: T=199 and T=200 for Patient A, and T=1099 and T=1100 for Patient B. The red bars mark the midpoint of the two selected time steps, highlighting that the patient was in circulatory failure during this period. Each timestep covers a 5-minute interval.}
\label{fig1} % Add a label for cross-referencing
\end{figure}

% \vspace*{-\baselineskip}
We observed that GradientSHAP, DeepLift, Integrated Gradients, Occlusion, Permutation, Ablation and their variants are not inherently optimized for time-varying multi-output models. This becomes particularly challenging in dynamic prediction tasks, where predictions are generated at every time step, resulting in extended attribution outputs. Specifically, these methods produce attributions of size \( T \times F \) for each time step \( T \) and class \( C \), which results in a full attribution set of \( T \times T \times F \times C \), where \( F \) is the number of features. This failure mode is illustrated in Figure \ref{fig1} for the GradientSHAP method, where we analyzed the ICU stay records of Patient A and Patient B (see methods) involving circulatory failure. Patient A stayed in the ICU for 7 days with three brief periods of elevated risk of circulatory failure. Patient B stayed in the ICU for 7 days with two brief periods of elevated risk of circulatory failure. 

\begin{figure}[H]
\begin{center}
\includegraphics[width=\linewidth]{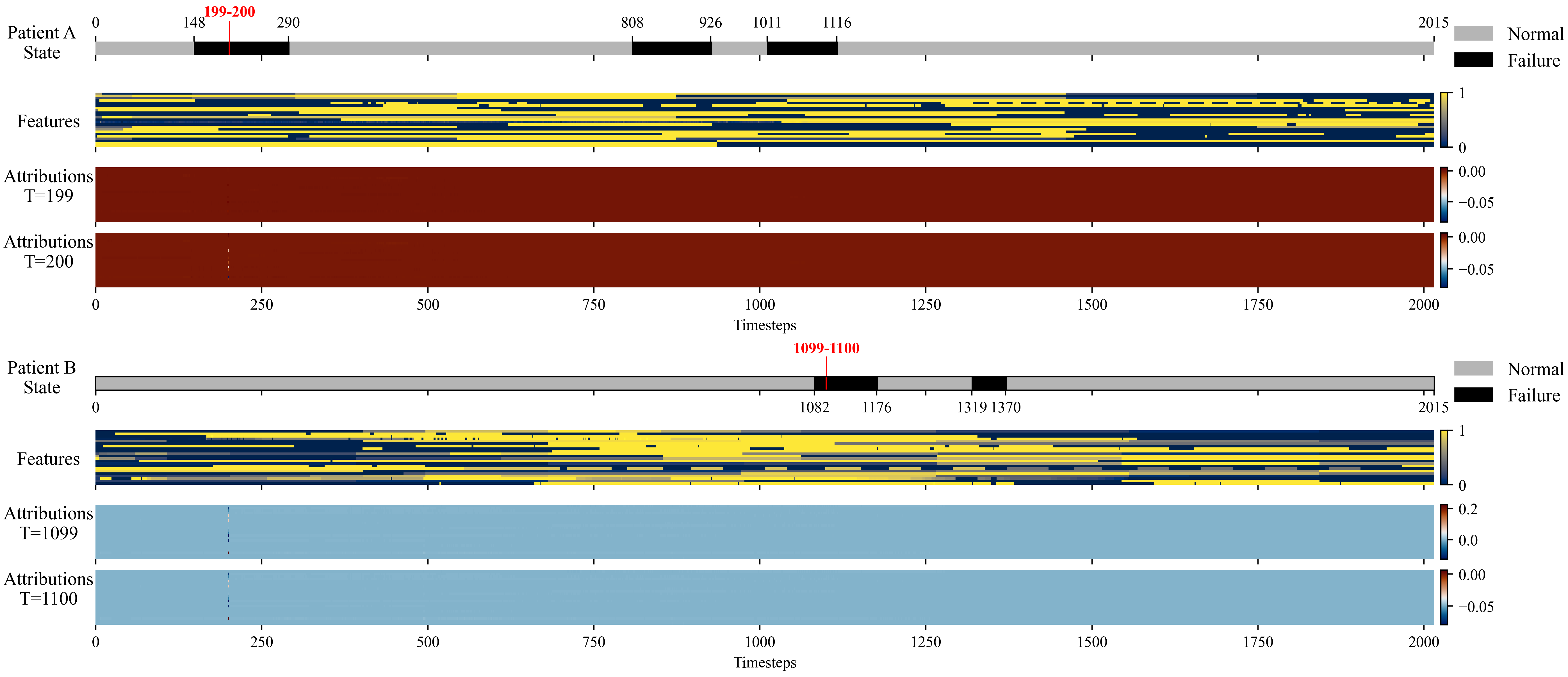}
\end{center}
\caption{Illustration of the timeline for Patient A and Patient B, from T=0 to T=2015, for the dynamic prediction task of circulatory failure. For each patient, the top bar indicates the patient's state over time, while the first heatmap displays normalized feature values across the timeline. Additionally, feature-time attribution heatmaps generated using the \textbf{Integrated Gradients (IG)} method are presented for specific time steps: T=199 and T=200 for Patient A, and T=1099 and T=1100 for Patient B. The red bars mark the midpoint of the two selected time steps, highlighting that the patient was in circulatory failure during this period.Each timestep covers a 5-minute interval.}

\label{fig2} % Add a label for cross-referencing
% \label{fig:appendix-1}
\end{figure}

\begin{figure}[H]
\begin{center}
\includegraphics[width=\linewidth]{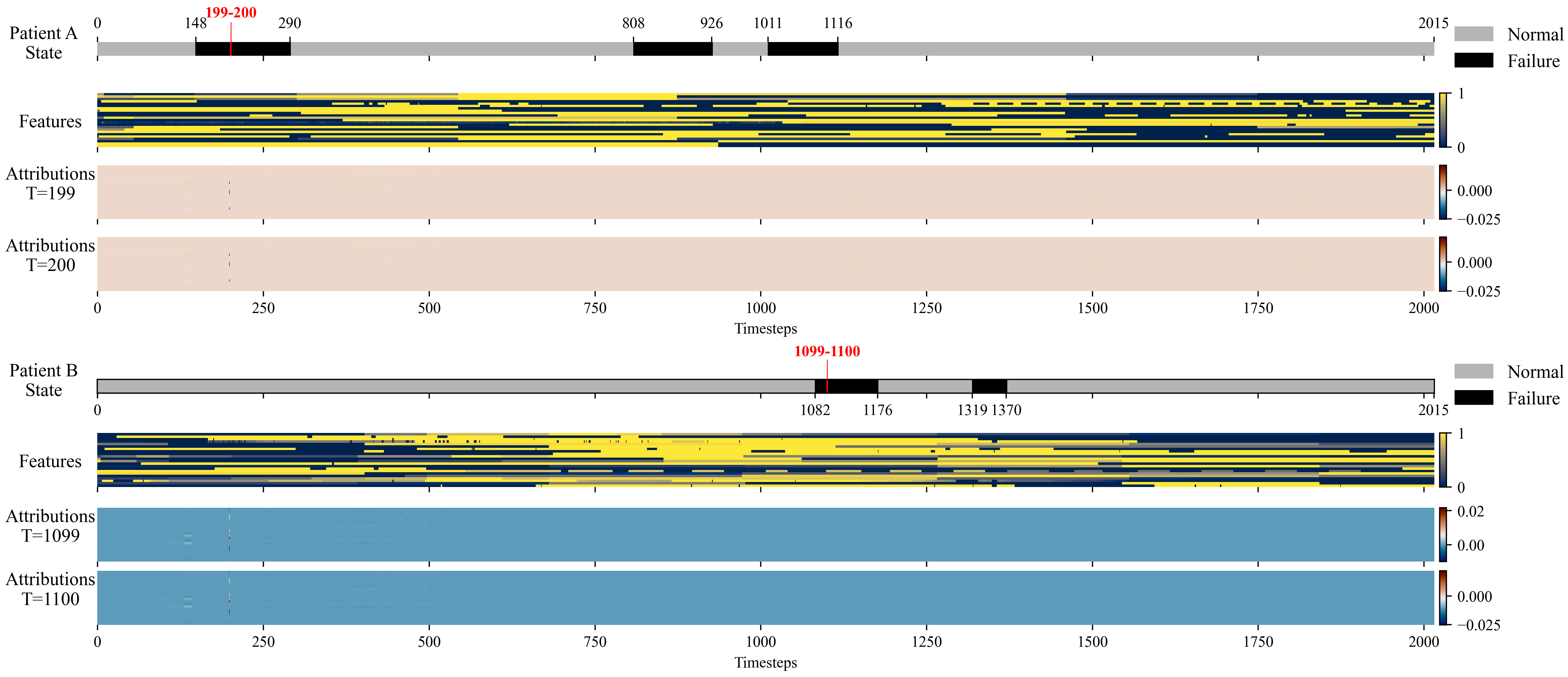}
\end{center}
\caption{Illustration of the timeline for Patient A and Patient B, from T=0 to T=2015, for the dynamic prediction task of circulatory failure. For each patient, the top bar indicates the patient's state over time, while the first heatmap displays normalized feature values across the timeline. Additionally, feature-time attribution heatmaps generated using the \textbf{DeepLift} method are presented for specific time steps: T=199 and T=200 for Patient A, and T=1099 and T=1100 for Patient B. The red bars mark the midpoint of the two selected time steps, highlighting that the patient was in circulatory failure during this period. Each timestep covers a 5-minute interval.}
\label{fig3} % Add a label for cross-referencing
% \label{fig2}
\end{figure}

\begin{figure}[H]
\begin{center}
\includegraphics[width=\linewidth]{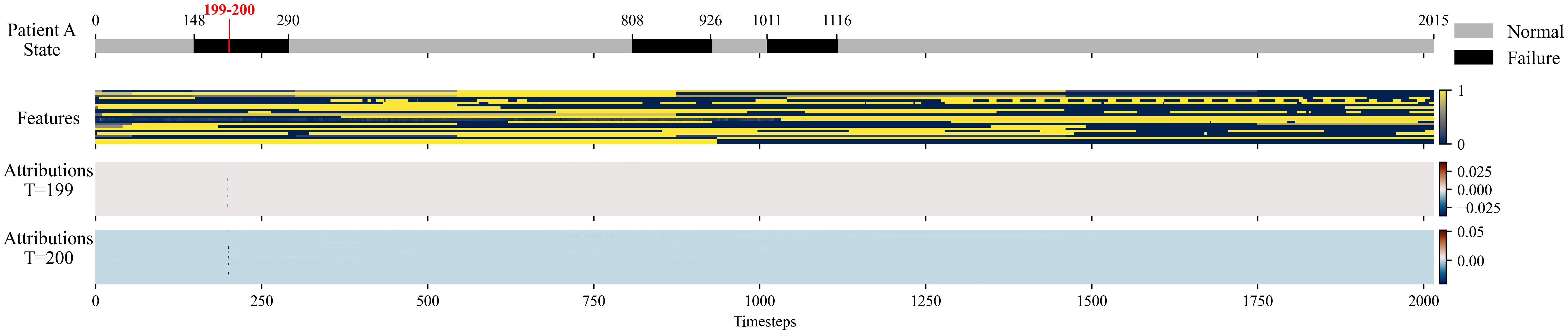}
\end{center}
\caption{Illustration of Patient A’s timeline from T=0 to T=2015 for the dynamic circulatory failure prediction task. The top bar represents Patient A's state over time, while the first heatmap displays normalized feature values. The two lower heatmaps depict feature-time attributions from the \textbf{DeepLiftSHAP} method at selected time steps (T=199 and T=200). The red bars indicate the midpoint of the selected time steps, highlighting that the patient was in circulatory failure during this period. Each timestep is 5 minutes. Patient B’s timeline and heatmaps omitted (space constraints).}
\label{fig4} % Add a label for cross-referencing
% \label{fig:appendix-3}
\end{figure}

\begin{figure}[H]
\begin{center}
\includegraphics[width=\linewidth]{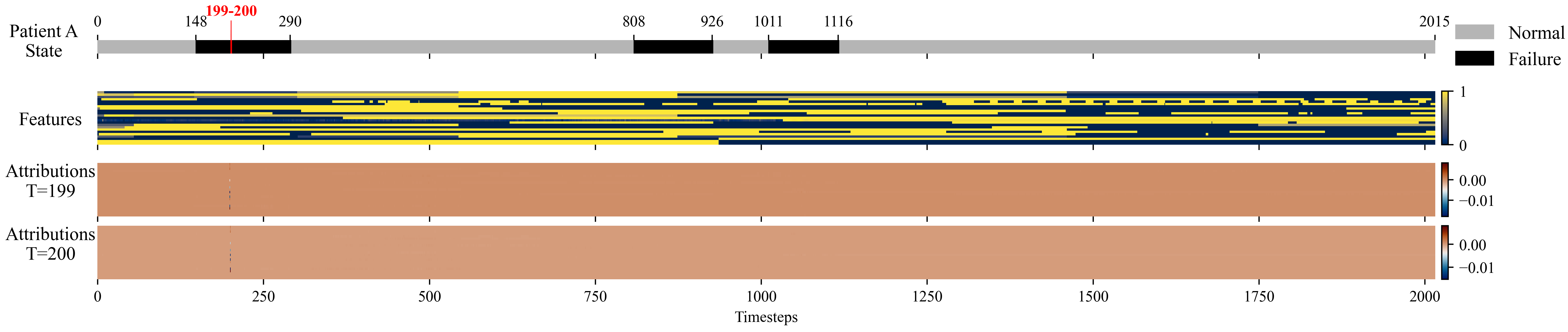}
\end{center}
\caption{Illustration of Patient A’s timeline from T=0 to T=2015 for the dynamic circulatory failure prediction task. The top bar represents Patient A's state over time, while the first heatmap displays normalized feature values. The two lower heatmaps depict feature-time attributions from the  \textbf{Sequential Integrated Gradients} method at selected time steps (T=199 and T=200). The red bars indicate the midpoint of the selected time steps, highlighting that the patient was in circulatory failure during this period. Each timestep is 5 minutes. Patient B’s timeline and heatmaps omitted.}
\label{fig5} % Add a label for cross-referencing
% \label{fig:appendix-4}
\end{figure}

\begin{figure}[H]
\begin{center}
\includegraphics[width=\linewidth]{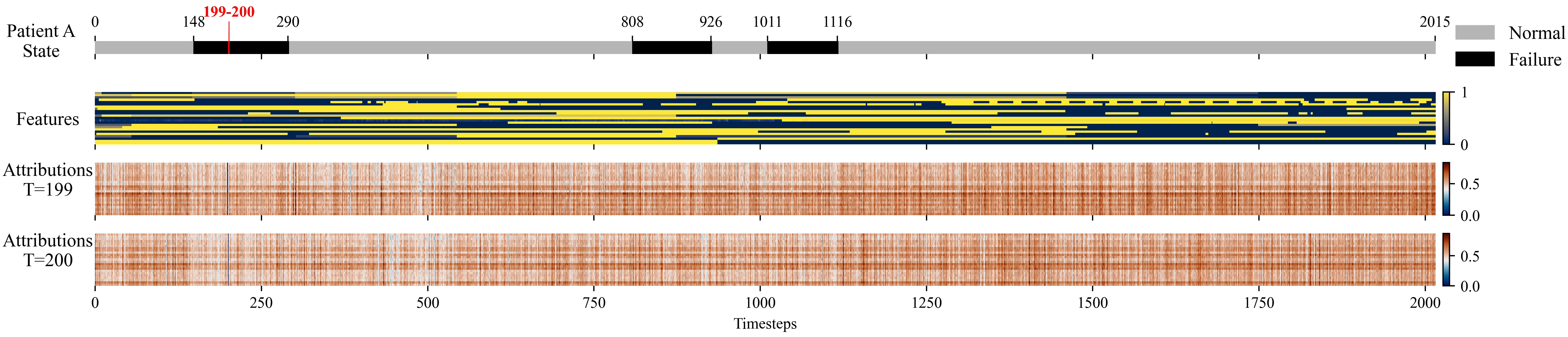}
\end{center}
\caption{Illustration of Patient A’s timeline from T=0 to T=2015 for the dynamic circulatory failure prediction task. The top bar represents Patient A's state over time, while the first heatmap displays normalized feature values. The two lower heatmaps depict feature-time attributions from the  \textbf{Feature Occlusion} method at selected time steps (T=199 and T=200). The red bars indicate the midpoint of the selected time steps, highlighting that the patient was in circulatory failure during this period. Each timestep is 5 minutes. Patient B’s timeline and heatmaps omitted (space constraints).}
\label{fig6} % Add a label for cross-referencing
% \label{fig:appendix-5}
\end{figure}

\begin{figure}[H]
\begin{center}
\includegraphics[width=\linewidth]{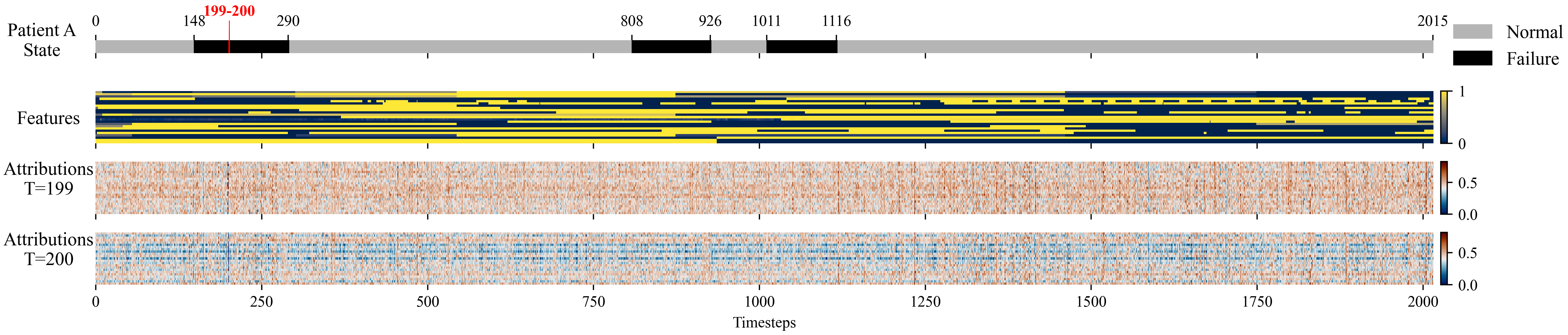}
\end{center}
\caption{Illustration of Patient A’s timeline from T=0 to T=2015 for the dynamic circulatory failure prediction task. The top bar represents Patient A's state over time, while the first heatmap displays normalized feature values. The two lower heatmaps depict feature-time attributions from the  \textbf{Augmented Feature Occlusion} method at selected time steps (T=199 and T=200). The red bars indicate the midpoint of the selected time steps, highlighting that the patient was in circulatory failure during this period. Each timestep is 5 minutes. Patient B’s timeline and heatmaps omitted.}
\label{fig7} % Add a label for cross-referencing
% \label{fig:appendix-6}
\end{figure}

\begin{figure}[H]
\begin{center}
% \centering
\includegraphics[width=0.93\linewidth]{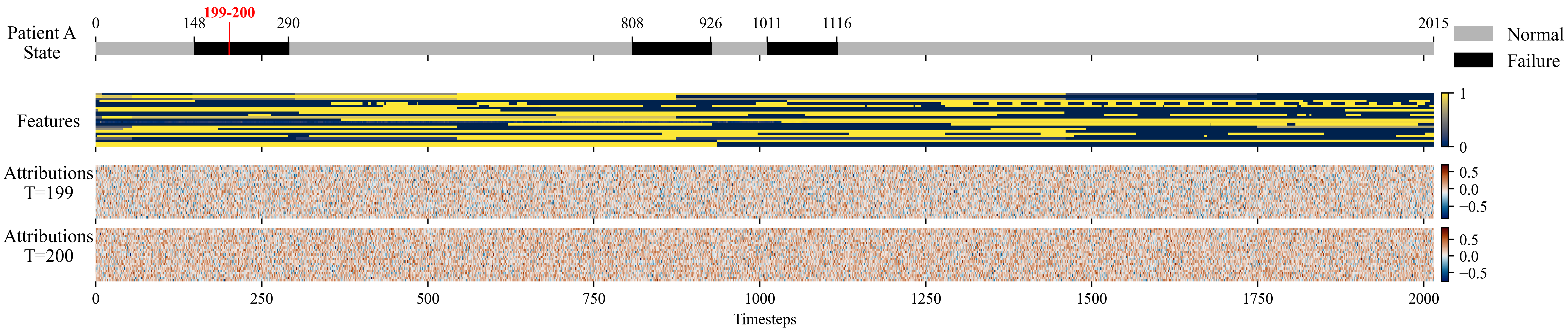}
\end{center}
\caption{Illustration of Patient A’s timeline from T=0 to T=2015 for the dynamic circulatory failure prediction task. The top bar represents Patient A's state over time, while the first heatmap displays normalized feature values. The two lower heatmaps depict feature-time attributions from the  \textbf{Feature Ablation} method at selected time steps (T=199 and T=200). The red bars indicate the midpoint of the selected time steps, highlighting that the patient was in circulatory failure during this period. Each timestep is 5 minutes. Patient B’s timeline and heatmaps omitted (space constraints).}
\label{fig8} % Add a label for cross-referencing
% \label{fig:appendix-7}
\end{figure}

\begin{figure}[H]
\begin{center}
% \centering
\includegraphics[width=\linewidth]{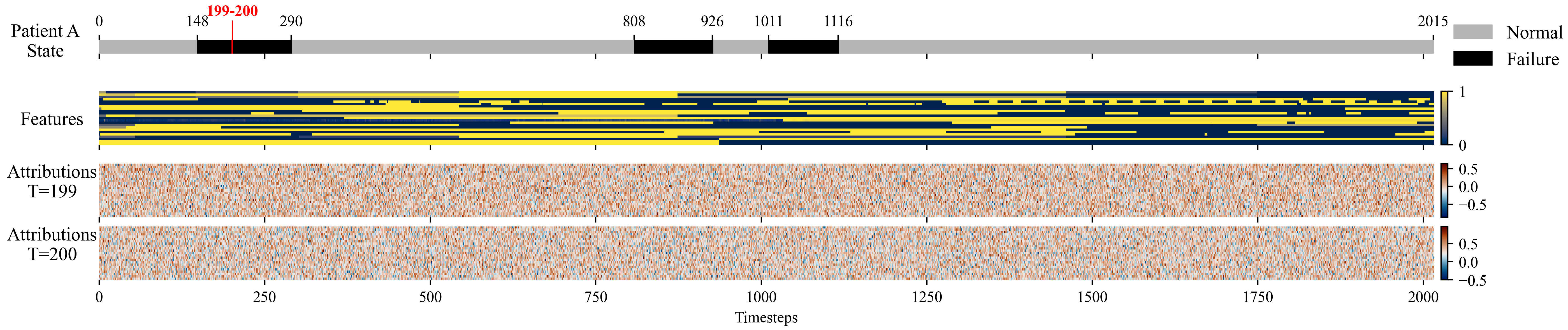}
\end{center}
\caption{Illustration of Patient A’s timeline from T=0 to T=2015 for the dynamic circulatory failure prediction task. The top bar represents Patient A's state over time, while the first heatmap displays normalized feature values. The two lower heatmaps depict feature-time attributions from the  \textbf{Feature Ablation} method at selected time steps (T=199 and T=200). The red bars indicate the midpoint of the selected time steps, highlighting that the patient was in circulatory failure during this period. Each timestep is 5 minutes. Patient B’s timeline and heatmaps omitted (space constraints).}
\label{fig9} % Add a label for cross-referencing
% \label{fig:appendix-8}
\end{figure}

To further validate these failure modes,  we selected two consecutive time steps (\( T =199, T=200\)) for Patient A and two consecutive time steps (\( T =1099, T=1100\)) for Patient B during a circulatory failure event and extracted the corresponding attribution maps for both records using all the gradients, SHAP, feature ablation methods, and their variants. For both records, the results showed that the attribution maps varied significantly across these adjacent time steps and contradicted the expectation that a short 5-minute interval would not result in substantial changes in the patient’s state, especially when the patient is already in a state of circulatory failure. Therefore, multi-dimensional attribution maps (size: \( T \times F \times C \)) generated for each time step \( T \) lack interpretable aggregation across the temporal dimension, obscuring sustained pathophysiological patterns. Similarly, other gradient, occlusion, ablation, and static permutation-based methods are plagued with the same failure modes as illustrated in Figures (\cref{fig2,fig3,fig4,fig5,fig6,fig7,fig8,fig9}), further highlighting the challenges in achieving temporal coherence in attributions.

Extending our analysis further, we observed that gradient and SHAP-based methods tend to assign the highest attribution to the current timestep while neglecting earlier timesteps, as illustrated in Figures (\cref{fig1,fig2,fig3,fig4,fig5,}. This results in empty attribution maps for past timesteps, limiting their usefulness in extracting insights about features that the model thought to be important before the onset of a circulatory failure event. The inability of these feature-time attribution maps to retain contributions from preceding time steps inadequately represents temporal interdependencies, hindering retrospective analysis of how predictive biomarkers evolve during the onset of an organ failure event. Ultimately, these almost empty attribution maps lack informativeness about the model's behavior, offering minimal insight into its decision-making process. In addition, multi-dimensional attribution maps with dimensions \( T \times T \times F \times C \), currently lack a clear method for combining information across the temporal dimension. For example, in Figure 1 for Patients A and B, each of the two selected timesteps produces an attribution map of the shape \( T \times F\) when extracted for the positive class (C=1). However, it is not clear whether these maps should be averaged, merged using a rolling window or combined by some other method. Without an effective aggregation strategy, it becomes difficult to identify sustained pathophysiological patterns before or close to the onset of circulatory failure. As a result, it is challenging to extract persistent temporal patterns that span multiple timesteps. For multi-dimensional attribution maps, it is even harder to determine how early indicators or gradual changes in the patient’s condition contribute to the eventual outcome. Moreover, interpretability heatmaps at a specific time \( T  \), often incorporate attribution scores influenced by future observations (such as timesteps \( T+1, T+2, ...  \), a common characteristic of feature occlusion, feature ablation, and feature permutation-based methods, as illustrated in Figures (\cref{fig7,fig8,fig9}). This observation violates clinical causality by allowing future information to influence interpretations of past events.

% It further leads to two sub-modes of failure: 

% \begin{enumerate}[label=\roman*., topsep=0pt, leftmargin=1.5em, labelwidth=1.2em, align=left]
%     % \textcolor{red}{
%     \item Temporal Aggregation: Multi-dimensional attribution maps (size: \( T \times F \times C \)) generated for each time step \( T \) lack interpretable aggregation across the temporal dimension, obscuring sustained pathophysiological patterns.
%     % }
%     % \textcolor{red}{
%     \item Temporal Causality: Interpretability heatmaps for time 
% \( T \) incorporate attribution scores influenced by future observations (e.g., \( T + 1\), \( T + 2\), ...), violating clinical causality.
% % }
%     \end{enumerate}

\setlength{\abovecaptionskip}{2pt}

Occlusion and augmented occlusion methods require selecting sliding window shapes as a hyperparameter in their formulation, which poses a challenge because the optimal time duration of feature activation leading to circulatory failure remains undetermined. Moreover, using a window that is too short not only results in an extremely high runtime but also introduces issues with the temporality of attribution scores. In addition, feature ablation and feature permutation operate on one feature-time point at a time, a strategy that is problematic for a densely sampled time series consisting of 465,696 feature-time point pairs (231×2016). Processing them individually is not feasible and disrupts the temporal continuity essential in a clinical context, as evidenced by attribution maps that resemble Gaussian noise. The limitations in temporal coherence are explored in detail as a failure mode for other methods in the next section.

\begin{figure}[H]
\begin{center}
\includegraphics[width=\linewidth]{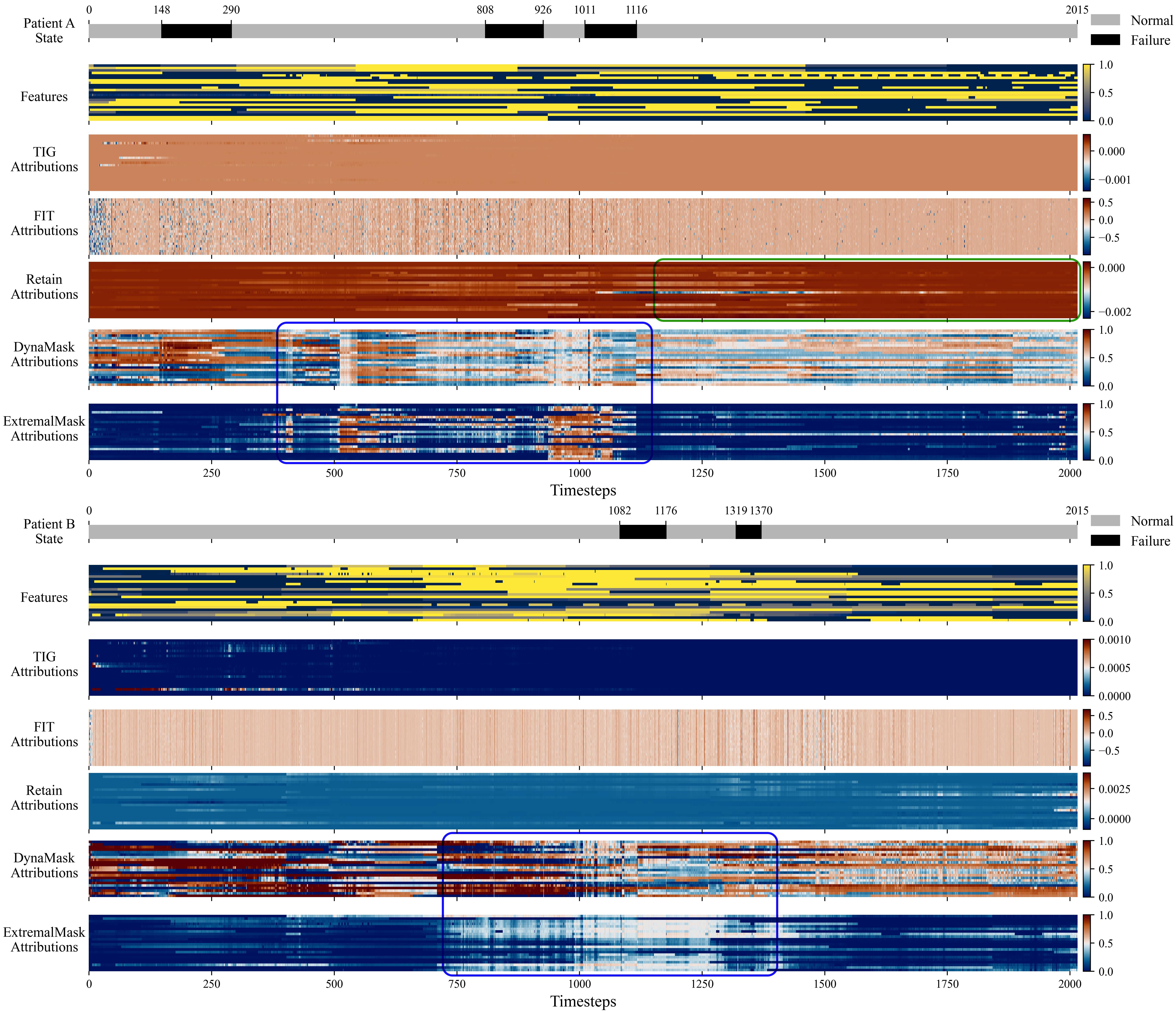}
\end{center}
\caption{Illustration of Patient A's timeline from T=0 to T=2015 for the dynamic circulatory failure prediction task. The top bar represents the patient's state over time, while the first heatmap displays normalized feature values. The next five heatmaps show attributions from \textbf{Temporal Integrated Gradients (TIG)}, \textbf{Feature Importance in Time (FIT)}, \textbf{RETAIN}, \textbf{DynaMask}, and \textbf{ExtremalMask}, respectively, across all time steps. Each timestep covers a 5-minute interval. For Patient B, the same set of visualizations is shown, providing a comparative view of feature attributions and patient state dynamics over time. Blue rectangles highlight regions where mask-based methods produce strong attributions prior to the failure event. In contrast, the green rectangle in the \textbf{RETAIN} heatmap illustrates a failure mode where nearly all features are attributed high importance after the onset, violating causal constraints.}
\label{fig10} % Add a label for cross-referencing
\end{figure}

\subsection{Failure Mode 2: Issues with Temporal Smoothness}
% \vspace*{-\baselineskip}
Interpretability methods specifically designed for time-series contexts, such as TIG and FIT, produce attributions that are not coherent in time. We observed that feature attributions for these methods show abrupt changes in feature importance that are inconsistent with the smooth trends typically observed in ICU time-series data. This behavior is illustrated in Figure \ref{fig10} for both TIG and FIT. For TIG, the feature-time attribution maps exhibit some temporal coherence, but the scores largely concentrate on just one or two features, and the absolute values across the temporal range remain fairly uniform. This suggests that there is no clear indication of which features are primarily responsible for triggering the onset of circulatory failure. For FIT, the temporal incoherence may stem from its dependence on pointwise Kullback-Leibler (KL) divergence to quantify predictive distributional shifts, which isolates individual time steps by contrasting observed features against counterfactuals where other features are masked. While theoretically rigorous, this framework risks overemphasizing transient perturbations in the predictive distribution rather than capturing the gradual, interdependent evolution of patient states. The RETAIN method, though, provides temporally coherent attributions that align with the sequence of patient data; the attribution values span only a narrow range, making it harder to distinguish subtle differences in feature-time importance. Our analysis shows that significant attributions tend to emerge only after failure events have already occurred, as illustrated in Figure \ref{fig10}. Additionally, RETAIN was originally designed to handle discretely segmented encounters, treating each hospital visit as an independent event rather than addressing continuous, high-resolution data from a single stay. This means it does not fully capture the fine-grained temporal dynamics inherent in prolonged patient monitoring.

\subsection{Potential Solutions: Leveraging Perturbation-Based Mask Methods}
% \vspace*{-\baselineskip}

Recently, DynaMask and ExtremalMask have emerged as mask-based perturbation methods for modeling time-series model interpretability. They learn masks over the input time series to quantify attribution scores for each feature-time pair, but they differ in how they perturb the masked regions. In DynaMask, the mask is optimized based on label consistency, sparsity, and temporal smoothness, while fixed local operations—such as Gaussian blurs or moving averages—are used to perturb the masked regions. Although this approach effectively maintains temporal continuity, its reliance on local perturbations limits the modeling of long-range dependencies common in clinical time series. In contrast, ExtremalMask jointly learns both a binary mask and a context-aware perturbation function that adapts to the input’s global temporal structure, thereby preserving overall temporal dynamics while highlighting salient regions.

For patients A and B, the attribution scores from DynaMask and ExtremalMask are illustrated in Figure \ref{fig10}. We observed that these methods achieve significantly lower variance in attributions across time steps, ensuring better temporal smoothness. Notably, the feature-time attributions produced by these mask-based methods score higher values before or close to the event onset, with attribution scores increasing as the prediction approaches the time of onset, highlighting their potential of capturing early trends in dynamic organ failure prediction tasks. Furthermore, additional constraints—including label consistency, temporal smoothness, minimal entropy, maximal informativeness, and in-distribution perturbation—can be incorporated during the optimization process, providing better control when developing interpretability methods. These techniques are completely model agnostic, and when combined with a causal model similar to ours, they can effectively pinpoint the features that are potentially responsible for predicting the onset of organ failure. We argue that the potential of learnable mask-based approaches is immense and they can offer consistent and meaningful interpretations for dynamic prediction tasks.

% \begin{figure}[H]
% \begin{center}
% % \centering
% \includegraphics[width=\linewidth]{XAI_10_RETAIN_Example.png}
% \end{center}
% \caption{Illustration of a patient’s timeline from T=0 to T=2015 for the dynamic circulatory failure prediction task. The top bar shows the patient's state. The first heatmap shows normalized feature values over time. The second heatmap shows attributions obtained from the \textbf{RETAIN} method for the whole timeline.}
% \label{fig11} % Add a label for cross-referencing
% \end{figure}

% \begin{figure}[h]
% \begin{center}
% \includegraphics[width=\linewidth]{XAI_08_D_E_Example.png}
% \end{center}
% \caption{Temporal Integrated Gradients (TIG) and Feature Importance in Time (FIT) Attributions}
% \label{fig2} % Add a label for cross-referencing
% \end{figure}

\section{Discussion}
% \vspace*{-\baselineskip}

In this work, we examine the failure modes of traditional time-series interpretability algorithms when applied to dynamic organ failure prediction, shedding light on critical limitations and providing potential solutions. While previous work by Srinivas \& Fleuret \citep{srinivas2020rethinking} demonstrated the limitations of gradient-based interpretability methods, their analyses did not address the unique challenges of dynamic time-series data. In our work, we specifically focus on dynamic organ failure prediction, contributing to the development and application of model-agnostic time series interpretability methods tailored for critical care applications. Our study qualitatively evaluated 14 algorithms on a dynamic circulatory failure prediction task, revealing two broad failure modes: I) Time-Varying Multi-Output Models and II) Challenges with maintaining Temporal Smoothness. When interpreting multi-dimensional attribution maps, as produced by Gradient, SHAP, Occlusion, Ablation, and Permutation methods, it is often difficult to determine how early pathophysiological indicators contribute to the onset of a circulatory failure event. Also, some methods may inadvertently violate clinical causality by allowing future observations to influence interpretations of earlier events. Moreover, although many techniques claim to be model-agnostic, they frequently rely on assumptions that break down in causal temporal contexts, thereby underscoring the need for truly causal interpretability methods. Methods, such as FIT, can overemphasize short-term changes instead of capturing the gradual, interconnected progression of patient conditions, breaking the temporal smoothness requirement. In contrast to traditional methods, mask-based model interpretability approaches offer a significant advantage for interpretability in time-series tasks. These methods ensure that causality is maintained and feature importance scores evolve smoothly across adjacent time steps by integrating temporal continuity and label consistency, which preserves the inherent dependencies in time series. We believe that learnable mask-based approaches show great promise in providing consistent and meaningful interpretations for dynamic prediction and similar tasks in critical care.

Our work focuses on interpretability as a tool for model development and debugging, rather than clinically relevant explanations for, say, bedside decision support. In this context, temporally coherent and causally grounded attributions support modelers in validating and refining dynamic prediction models. Future research could examine how temporally coherent attributions perform in practical settings, such as through clinician-in-the-loop evaluations. These efforts may help shift model interpretability methods from internal diagnostic tools to practical applications that enhance both understanding of model behavior and the extraction of actionable clinical insights.

\section{Code Availability}
% \vspace*{-\baselineskip}
 The code to reproduce the results is available at: \\
 \href{https://github.com/xinformatics/txaifailuremodes}{\url{https://github.com/xinformatics/txaifailuremodes}}. 

% \subsection{Dataset description}

% \subsection{Model Evaluation}

% \subsection{Comparison of model interpretability methods}

% \bibliography{amia}
% \bibliographystyle{vancouver}
{\fontsize{10pt}{10pt}\selectfont
\bibliographystyle{vancouver}
\bibliography{amia}
}
% {\footnotesize
% % \setlength{\baselineskip}{4pt}  % adjust line height to match smaller font
% % \linespread{0.4}\selectfont  % try 0.85–0.95 for tighter spacing
% \bibliographystyle{vancouver}
% \bibliography{amia}
% }
\end{document}